%% file: ieeeconf-main.tex
\documentclass[a4paper, 10pt, conference]{ieeeconf/IEEEtran}  

\IEEEoverridecommandlockouts                              

\usepackage{graphicx} 
\usepackage{epsfig} 
\usepackage{mathptmx} 
\usepackage{times} 
\usepackage{amsmath} 
\usepackage{amssymb}  
\usepackage{bbm}

\usepackage{hyperref}
\hypersetup{
  colorlinks=false, 
  bookmarks=true, 
  bookmarksnumbered=true,
  pdfborder={0 0 0},
  bookmarkstype=toc
}
\usepackage{booktabs}
\usepackage[caption=false, font=footnotesize]{subfig}
\usepackage[%
backend=biber,
url=true,
maxbibnames=2,
minbibnames=1,
bibstyle=ieee,
]{biblatex}
\AtEveryBibitem{
  \clearfield{doi}
  \clearfield{isbn}
  \clearfield{issn}
  \clearfield{month}
  \ifentrytype{inproceedings}{
    \clearfield{arxivId}
    \clearfield{archivePrefix}
    \clearfield{eprint}
    \clearfield{url}
  }{}
  \ifentrytype{article}{
    \clearfield{arxivId}
    \clearfield{archivePrefix}
    \clearfield{eprint}
    \clearfield{url}
  }{}
  \ifentrytype{report}{
  }{}
}
\renewbibmacro{in:}{%
  \ifentrytype{inproceedings}{%
    \setunit{}
    \addperiod\addspace In \textit{Proc.\ of the}}%
  {\printtext{\bibstring{in}\intitlepunct}}
}
\DeclareSourcemap{
  \maps{
    \map{ 
      \step[fieldsource=url,
      match=\regexp{\{\\\_\}|\{\_\}|\\\_},
      replace=\regexp{\_}]
    }
    \map{ 
      \step[fieldsource=url,
      match=\regexp{\{\$\\sim\$\}|\{\~\}|\$\\sim\$},
      replace=\regexp{\~}]
    }
    \map{ 
      \step[fieldsource=url,
      match=\regexp{\{\\\x{26}\}},
      replace=\regexp{\x{26}}]
    }
  }
}
\title{\LARGE \bf
Online Refinement of a Scene Recognition Model for Mobile Robots 
by Observing Human's Interaction with Environments
}

\author{
Shigemichi Matsuzaki$^{1}$,
Hiroaki Masuzawa$^{1}$,
and Jun Miura$^{1}$ 
\thanks{
The first author is supported in part through the Leading Graduate School Program, 
“Innovative program for training brain-science-information-architects by analysis of massive quantities of highly technical information about the brain,”
by the Ministry of Education, Culture, Sports, Science and Technology, Japan.
}
\thanks{$^{1}$Department of Computer Science and Engineering,
Toyohashi University of Technology, Japan.
{\tt\small matsuzaki@aisl.cs.tut.ac.jp, 
\{masuzawa.hiroaki.vl, jun.miura\}@tut.jp }
}%
}

\begin{document}

\maketitle
\thispagestyle{empty}
\pagestyle{empty}


\begin{abstract}

	This paper describes a method of online refinement of
	a scene recognition model
	for robot navigation considering \textit{traversable plants},
	flexible plant parts which a robot can push aside while moving.
	In scene recognition systems that consider traversable plants
	growing out to the paths,
	misclassification may lead the robot to getting stuck
	due to the traversable plants recognized as obstacles.
	Yet, misclassification is inevitable 
	in any estimation methods.
	In this work, we propose a framework 
	that allows for refining a semantic segmentation model
	on the fly during the robot's operation. 
	We introduce a few-shot segmentation based on weight imprinting
	for online model refinement without fine-tuning.
	Training data are collected via observation of
	a human's interaction with the plant parts.
	We propose novel robust weight imprinting to mitigate the effect 
	of noise included in the masks generated by the interaction.
	The proposed method was evaluated through experiments
	using real-world data and shown to
	outperform an ordinary weight imprinting
	and provide competitive results to 
	fine-tuning with model distillation
	while requiring less computational cost.
\end{abstract}

\input{sections/introduction}

\input{sections/related_work}

\input{sections/proposed_method}

\input{sections/experiments}

\input{sections/conclusions}



\printbibliography

\end{document}

%% file: sections/introduction.tex
\section{INTRODUCTION}

Navigation in plant-rich environments is
a challenging problem due to the presence of
plant parts covering the paths. 
In scene recognition using range sensors,
which are widely adopted in conventional mobile robots,
those plant parts are recognized as obstacles
blocking the paths.
Some of such plant parts are, however, flexible and
a robot can move through them by pushing them aside.
We call such plants \textit{traversable plants} in this work.

\begin{figure}[tb]
	\centering
	\includegraphics[width=0.80\hsize]{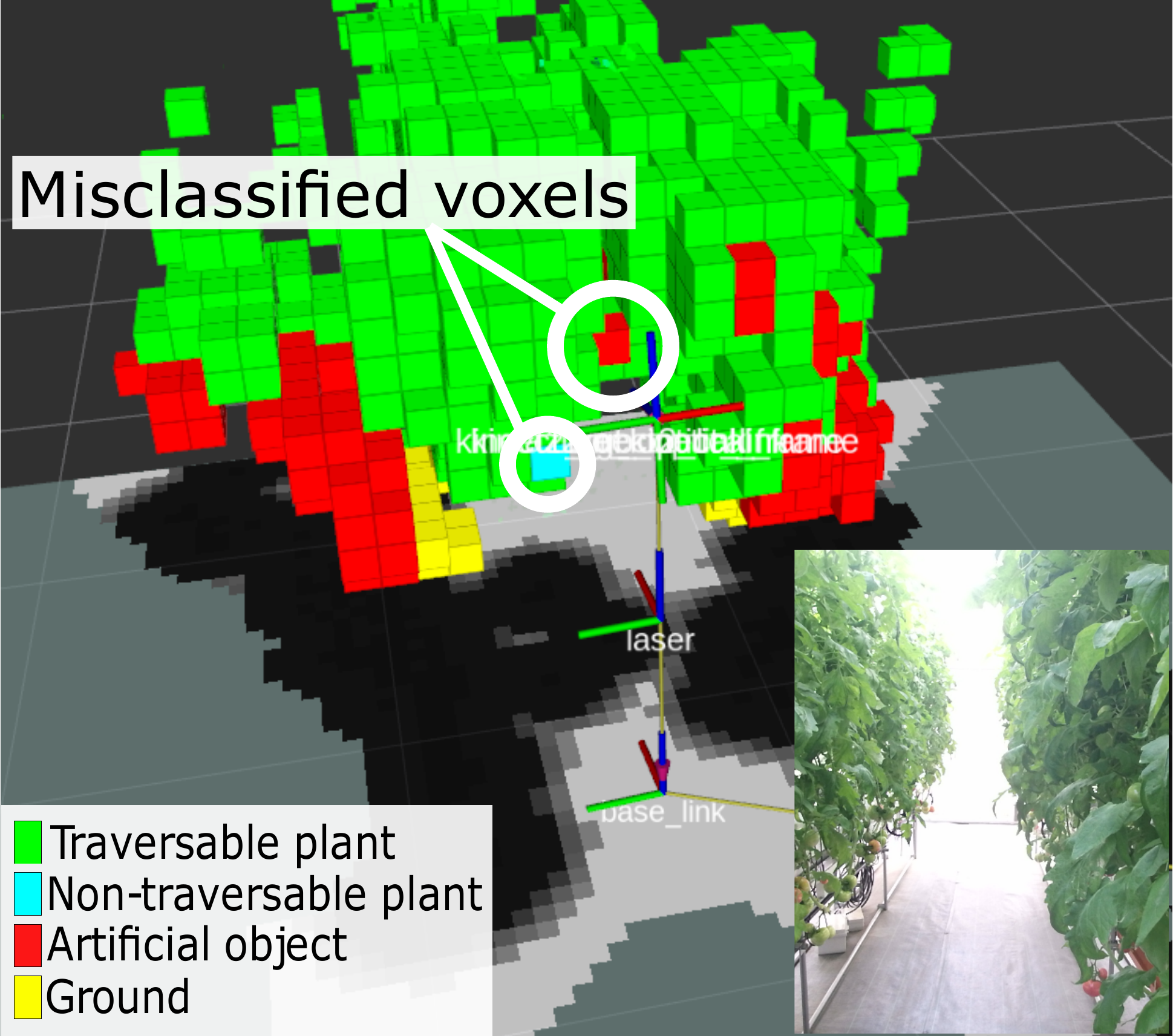}
	\caption{Misclassification in the conventional system}
	\label{fig:misclassification}
\end{figure}

Presence of traversable plants in the context of
robot navigation has not been dealt in many studies.
A pioneering work by Kim et al. \cite{Kim2006}
estimated the traversability of the regions in an image
using color and geometric features.
The authors' previous work \cite{Matsuzaki2022} proposed a novel
deep neural network (DNN) based on
semantic segmentation for object traversability
estimation with a manual annotation-free training method.
Although we demonstrated the effectiveness of the method
through navigation experiments in a plant-rich greenhouse,
misclassifications were a critical problem.
Fig. \ref{fig:misclassification} shows an example of a case
where some voxels of traversable plants are
misclassified as other obstacles and the path is considered blocked.
When such a misclassification often occurs,
the robot easily gets stuck.
Such misclassifications are inevitable in
scene recognition methods including DNNs.
A method to deal with them on the fly is,
therefore, needed
for practical uses of such a navigation system.

To this end, we are developing a method for
online refinement of the recognition model
exploiting a human's interaction with
the misclassified traversable plant regions
during the actual operation of mobile robots.
Ideally, the robot should indicate possibly misclassified regions
to the human with some interfaces to acquire data efficiently.
In this work, however,
we consider a situation where
the human randomly touches the plant parts.
We propose a framework for refining
the semantic segmentation model
making use of the information acquired through
such interaction.
The proposed method will be a building block of
more advanced system for traversable plant recognition
with a function for online refinement
of the recognition model. 

In the proposed method, a human first touches 
regions of plants in the 3D space
to provide the robot with information to correct the recognition of plants.
The robot observes the interaction
using a human pose detector such as OpenPose \cite{Cao2021}
to identify plant regions indicated by the human.
Masks of the interacted regions in images
are generated for the model refinement.
The scene recognition model is then refined using
the mask so that next time the robot encounters similar scenes,
the robot recognizes the scenes correctly.
More specifically, we introduce a few-shot learning method
based on weight imprinting \cite{Brown2018}.
Since the masks of interacted regions are generated
using depth images from an RGB-D sensor,
we add a modification to the process of weight imprinting
to deal with noise included in the masks.


The contributions of this work are as follows:
\begin{enumerate}
	\item A novel framework of online refinement of a scene recognition model
	      exploiting data acquired during the robot's operation
	      using a human's interaction with the environment.
	\item A novel weight imprinting-based few-shot segmentation
	      that is robust to noise in the masks.
\end{enumerate}


%% file: sections/related_work.tex
\section{RELATED WORK}

\subsection{Traversability estimation}

Traversability estimation is an essential task in
navigation of mobile robots.
Navigation in unstructured environments 
requires traversability estimation based on the terrain condition
such as steepness and roughness.
Major approaches include
terrain analysis using range sensors
\cite{Bogoslavskyi2013,,Martinez2020a,Zhou2021}
and image-based estimation of terrain semantics
\cite{Matsuzaki2018b,Maturana2018,Hosseinpoor2021}.

Some methods have been introduced for analyzing
the traversability of flexible objects such as plants.
Kim et al. \cite{Kim2006}
trained a classifier with hand-crafted color and geometric features
using a robot's experience of traversals.
We presented an image-based deep neural network for
object traversability estimation and its manual annotation-free
training method \cite{Matsuzaki2022}.
In \cite{Matsuzaki2022}, we reported some failure cases of navigation
due to misclassifications. 
Such machine learning-based methods
inevitably suffer from misclassifications.
To the best of our knowledge,
there has been no method proposed to
deal with the problem on the fly during the operation
in the context of robot navigation.

\subsection{Image segmentation via interaction with objects}

In the field of robotics,
many studies have proposed to leverage the robots'
active interaction with objects
to achieve information, which is not available in
passive observation
\cite{Li2011,Katz2013,Novkovic2020}.
Some other studies have also tackled problems
of learning an object segmentation model using
a human's interaction with the target objects
as supervisory signals.
Arsenio et al. \cite{Arsenio2003b} proposed an
object segmentation method through robot's own actions,
actions of a person with a wearable system,
or perceiving demonstrations of a human.
Kim et al. \cite{Kim2010b} proposed a method for
identification of objects to segment through a human's gesture.
Azagra et al. \cite{Azagra2020} also proposed a method to
indicate objects to learn to the robot via a human's gesture.

Existing work is focused on segmentation of
rigid body objects.
In contrast, our purpose is to identify
plant regions especially flexible parts such as
leaves and branches and to refine
misclassifications of the recognition model on the fly.

\subsection{Semantic segmentation with unseen classes}

Adapting a model to unseen classes of objects is
an actively studied task.
Tasks that involves such objective include
incremental learning, few-shot learning, continual learning,
lifelong learning etc. \cite{Michieli2021a}.

Incremental learning is a task to adapt a pre-trained model to
newly seen data while maintaining previously learned knowledge
avoiding catastrophic forgetting.
Knowledge distillation \cite{Hinton2015}, or model distillation,
is a widely used approach to the task
\cite{Michieli2021a,Cermelli2020,Klingner2020}.
Although the task tackled in the aforementioned studies
is relevant to ours,
those methods requires heavy computation such as backpropagation,
which is not suitable given limitation of computational resources
available on mobile robots.


Few-shot learning is one of machine learning paradigms
where limited new samples with ground truth labels (\textit{support set})
are given for reference to make inference on
data with object classes unseen in pre-training (\textit{query set}).
In few-shot segmentation,
a support set with ground truth mask is exploited to
aggregate features and
the aggregated features are used to produce classification weights for
segmentation of a query image \cite{Shaban2017,Nguyen2019,Li2021a}.

Qi et al. \cite{Brown2018} proposed a simple method
for few-shot image classification coined weight imprinting,
where normalized features of a target object
are directly used as weights of the classifier.
They claim that weight imprinting performs well even without fine-tuning.
Based on the weight imprinting approach, Siam et al. \cite{Siam2019a}
proposed the Adaptive Masked Proxies (AMP) for effective feature aggregation
specifically designed for few-shot image segmentation.
Our online learning is inspired by \cite{Brown2018}
and \cite{Siam2019a} for their simplicity.


%% file: sections/proposed_method.tex
\section{Review of the base methods}

\subsection{Traversability estimation}

Our previously proposed scene recognition model
considering traversable plants \cite{Matsuzaki2022}
consists of two modules;
Semantic Segmentation Module (SSM) and Traversability Estimation Module (TEM).
SSM provides information of general object classes,
namely \textit{plant}, \textit{artificial object}, and \textit{ground}.
TEM is a pixel-wise class-agnostic binary classifier
that estimates the traversability of each pixel
taking a feature map from SSM as input.

The estimation of the model is projected to the 3D space
using a depth image registered to the RGB image.
The 3D space is then divided into voxels with a fixed size
and the object probability and the traversability are
calculated for each voxel.
The estimation in each frame temporally fused via Bayesian update.

In the method,
misclassification of voxels led the robot
to getting stuck (see Fig. \ref{fig:misclassification}).
We reported multiple failures of
navigation due to misclassifications in \cite{Matsuzaki2022}.
The purpose of the present work is to amend the problem
by enabling online refinement of the recognition model
during the robot operation.
Because the main cause of voxel misclassification observed
in \cite{Matsuzaki2022} was
incorrect prediction of object classes
by SSM, we focus on refinement of
SSM in this work.

\subsection{Weight imprinting}
\label{sec:weight_imprinting_explanation}

Weight imprinting \cite{Brown2018}
was first proposed as a method
for few-shot image classification.
Given an embedding extractor
$\phi: \mathbb{R}^{N}\rightarrow\mathbb{R}^{D}$ which
maps an input $x \in\mathbb{R}^{N}$ to a $D$-dimensional
vector $\phi(x)\in\mathbb{R}^{D} $, and a softmax classifier
$f: \mathbb{R}^{D}\rightarrow\mathbb{R}^{C}$
which produces probability values for $C$ classes,
a probability for class $j$ is given as follows:
\begin{equation}
	f_j(\phi(x)) = \frac{\exp{(\mathbf{W}_j^T\phi(x))}}{\sum_c^{C}\exp{(w_{c}^T\phi(x))}},
	\label{eq:original_weight_imprinting}
\end{equation}
where $\mathbf{W}_j$ denotes the weight vector for class $j$.
Here, both the embedding $\phi(x)$ and the weight vector
$w_j$ are assumed to be L2-normalized.
Then, the main computation in eq. \eqref{eq:original_weight_imprinting}
is taking the inner product,
or equivalently the cosine similarity between
$\phi(x)$ and $w_j$.
Viewing each weight as a template vector
of the corresponding class,
the authors argued that the prediction
is equivalent to finding the template
closest to the embedding $\phi(x)$
based on the cosine similarity in the embedding space.
Based on the observation,
they proposed to use the embedding for
input data of a novel class
as the classification weight of the class.

Siam et al. \cite{Siam2019a} applied weight imprinting to
few-shot semantic segmentation.
Given the intermediate feature map
$\mathbf{x}_i\in\mathbb{R}^{H\times W\times D}$,
with the height $H$ and the width $W$,
and a corresponding binary mask $M_i\in\{0,1\}^{H\times W}$,
they aggregate embeddings over a given mask
via simple averaging followed by L2 normalization (masked average pooling: MAP):
\begin{equation}
	\begin{aligned}
		\label{eq:x_mean}
		\mathbf{x}_{MAP}       & =  \frac{\sum^{N}_{i=1}\sum^{H}_{h=1}\sum^{W}_{w=1}M^{(h,w)}\mathbf{x}^{(h, w)}_i }{\sum^{N}_{i=1}\sum^{H}_{h=1}\sum^{W}_{w=1}M^{(h,w)}} \\
		\hat{\mathbf{x}}_{MAP} & =  \frac{\mathbf{x}_{MAP}}{\|\mathbf{x}_{MAP}\|_2},
	\end{aligned}
\end{equation}
where $\left(h, w\right)$ denotes the coordinate of the image pixel.
The normalized average embedding $\hat{\mathbf{x}}_{MAP}$ is
used as classification weights of the novel class.
In \cite{Siam2019a},
they proposed to apply it to multiple layers to
exploit multi-resolution information, which is, in general,
considered important for semantic segmentation.
In our method, however, we apply weight imprinting
only on the last classification layer
for simplicity of implementation.

\section{Online model refinement}


Our aim is to refine the scene recognition model on the fly
with data collected during the robot's operation.
General fine-tuning is not suitable
due to limitations of computational resources and
of the variety of image data
acquired during the operation.
We, therefore, adopt a few-shot segmentation method
based on weight imprinting \cite{Brown2018},
which does not require fine-tuning
with costly backpropagation.

\subsection{Problem setting}

We assume that the scene recognition model $f(I)$
is pre-trained with a large amount of data
before deploying the robot in a target environment
following \cite{Matsuzaki2022} with $C$ classes.
We then assume to acquire a few pairs of
an RGB image and a corresponding binary mask
$S=\{(I_i, M_i)\}^{N}_{i=1}$,
where N denotes the number of the pairs.
The value 1 in the binary mask $M_i \in\{0, 1\}^{H\times W}$
indicates a pixel that should be learned as \textit{plant}
and the mask is acquired through interaction
by the human with the environment
during the robot operation.
The process of generating the mask is
described in the next subsection.

Our purpose is to update the model on the fly
with the image-mask pairs $S$
so that the model predicts the plant parts more accurately.
As mentioned above, we formulate the problem of
the online learning as a few-shot learning problem
where the objects of the masked regions are learned
as a novel $C+1$th class unseen in the pre-training.

\subsection{Data collection}
\begin{figure}[tb]
	\centering
	\includegraphics[width=0.7\hsize]{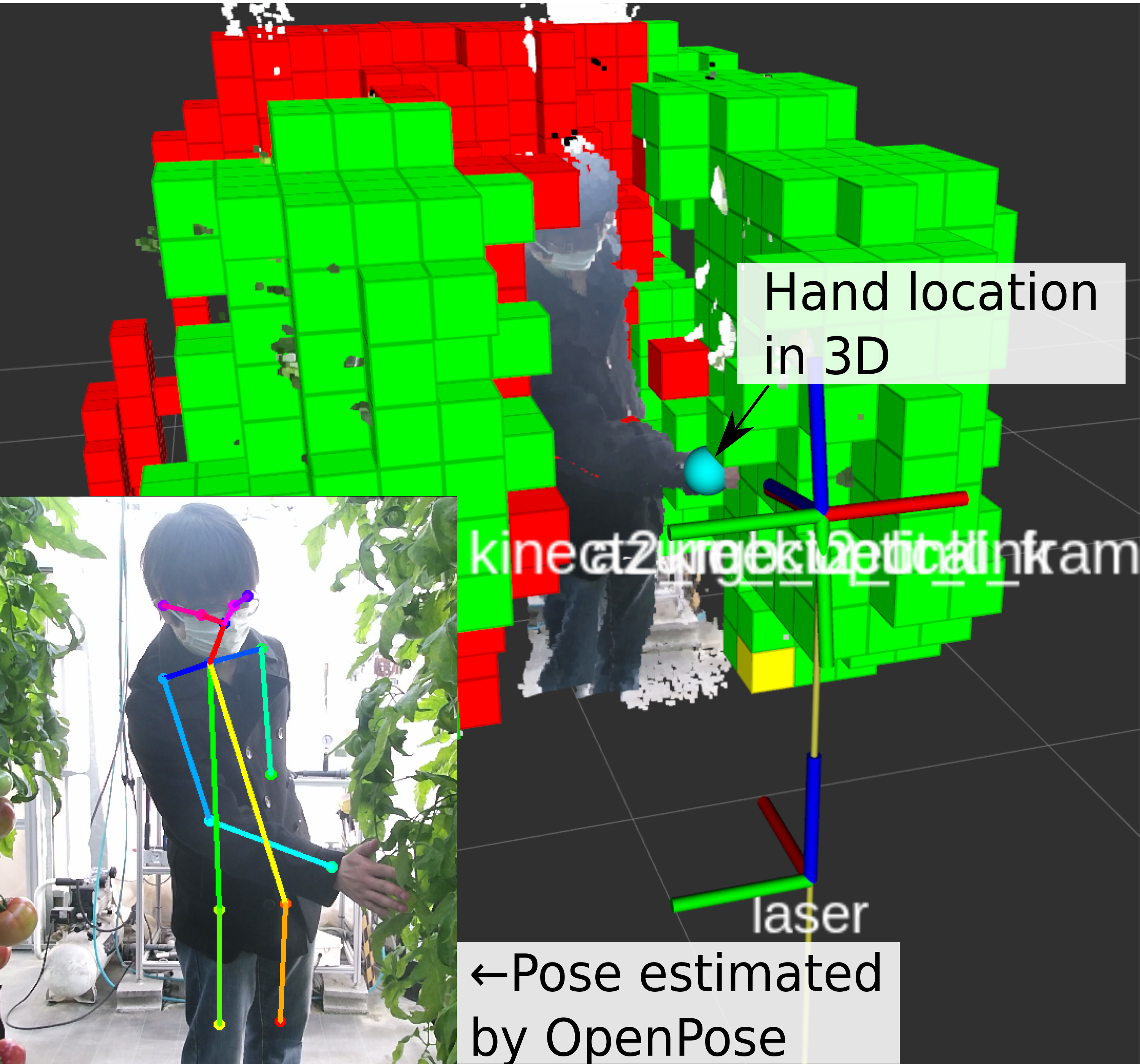}
	\vspace{-5pt}
	\caption{A human interacting with the plant parts}
	\label{fig:interaction}
\end{figure}


For few-shot learning, a pair of an input image $I$ and
a corresponding mask $M$
is generated based on the human's interaction.
The interaction mask indicates image regions that should be
classified as \textit{plant}.

\subsubsection{Labeling plant regions via a human's interaction}

First, a human interacts with the regions of
plant parts.
A robot observes the human's interaction using an RGB-D camera.
The human joints are recognized by OpenPose \cite{Cao2021}.
The coordinate of the right hand is projected into the 3D space
using the depth.
Voxels with which the human interacted are identified by
searching for voxels within a sphere with a radius of 5 [cm]
centered at the projected hand coordinate.

\subsubsection{Generating a training mask}

\begin{figure}[tb]
	\centering
	\includegraphics[width=\hsize]{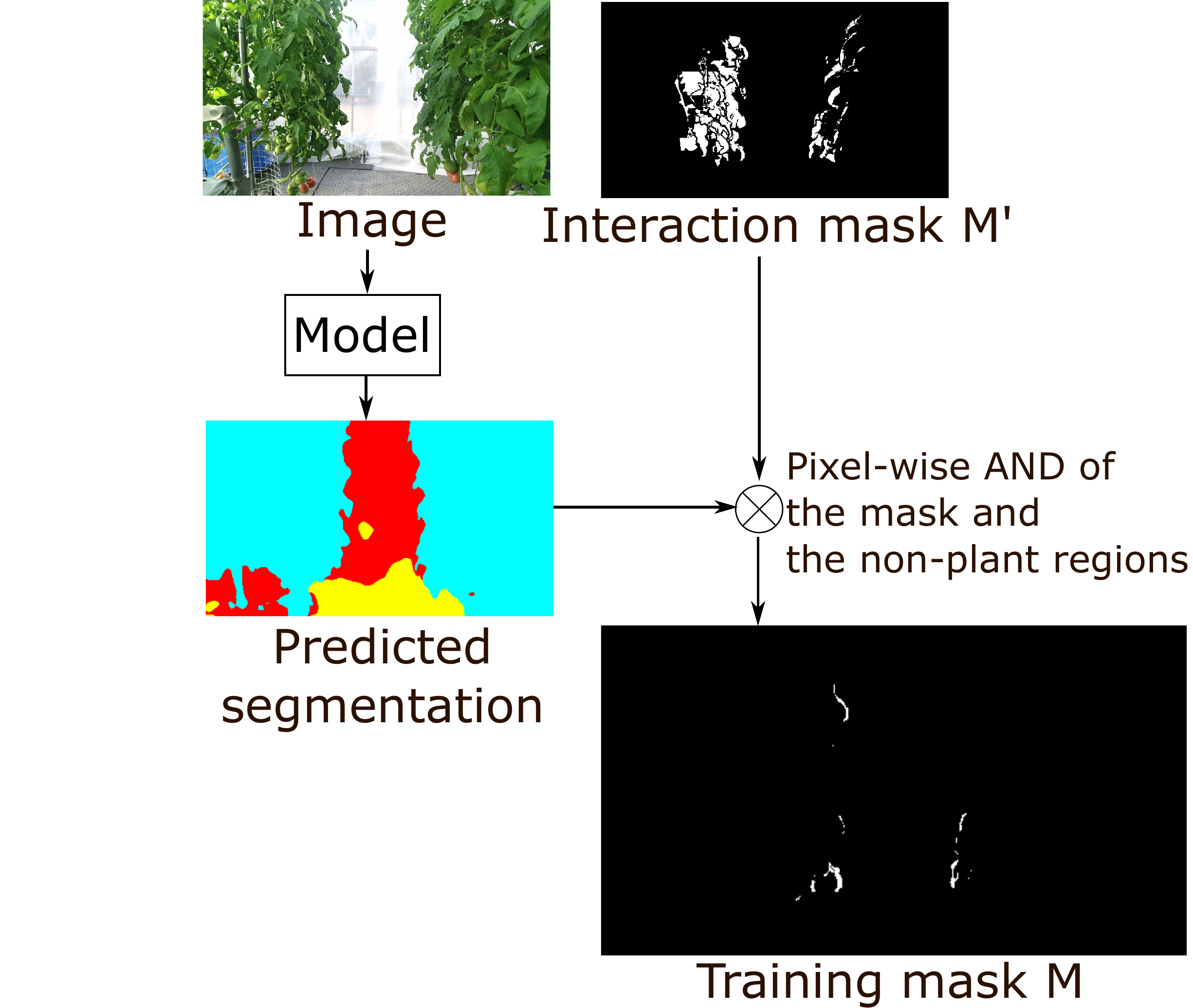}
	\vspace{-15pt}
	\caption{The process of generating a training mask from
		a pair of a camera image and an interaction mask}
	\label{fig:support_example}
\end{figure}

After the map labeling,
a mask $M$ for training is generated by
the process as follows.
\begin{enumerate}
	\item Get a pair of RGB and depth images.
	\item For each pixel of the depth image,
	      project it in the 3D space and search for a voxel
	      that the pixel belongs to.
	\item Label the pixel as 1 if the corresponding voxel has been
	      interacted, and otherwise 0.
	\item Repeat i) to iii) for five consecutive frames
	      and take pixel-wise OR of them
	      to deal with the depth noise.
	      We call the resulting mask
	      an \textit{interaction mask}
	      denoted as $M^{\prime}$.
	\item Predict object labels on the RGB image $I$
	      using the network and set pixels
	      of $M^{\prime}$ to 0 if the predicted object label of the corresponding pixels
	      are other than \textit{plant} (see Fig. \ref{fig:support_example}).
\end{enumerate}
We call the resulting mask $M$ a \textit{training mask}.
Here we assume that the interacted regions
in $M^{\prime}$ are dominated by plants.
The training mask $M$, therefore, indicates
false negative plant regions.

\subsection{Network architecture and the loss function}
\label{sec:network_architecture}

We adopt the network architecture proposed in \cite{Matsuzaki2022}
as a base network.
For the entire architecture of
the original network, readers are referred to
\cite{Matsuzaki2022}, Fig. 8.
Although we adopt the architecture in this work,
note that the proposed method is not dependent on a specific
network architecture.
To apply weight imprinting-based few-shot learning,
we make two modifications on the network architecture for
semantic segmentation and the loss function.

First, we apply L2 normalization on the weight vectors
of the final classification layer
$\{\mathbf{W}_j\}^{C}_{j=1}$, which is a $1\times 1$ convolution,
and the features
$\{\mathbf{x}^{(h,w)}_i\}^{N}_{i=1}$,
which is a feature vector in pixel $(h, w)$
of an intermediate feature map
yielded by forwarding an image $I_i$ through the model $f$.
The score of the features for each class is
then evaluated as a cosine similarity
between the features and the weight vector of the class
\cite{Brown2018}.

Second, to formulate the softmax loss
on the normalized data in a more natural way and
also to encourage the model to learn discriminative feature
representations, we introduce
the Additive Angular Margin Loss (ArcFace) \cite{Deng2021}
defined as follows:
\begin{equation}
	L_{arc}=-\sum^N_{i=1}\log{\frac{e^{s\cos({\theta_{y_i}+m})}}{e^{s\cos({\theta_{y_i}+m})}+\sum^C_{j=1,j\neq y_i}e^{s\cos{\theta_j}}}},
\end{equation}
where $\cos{\theta_j}=\mathbf{W}_j^T\mathbf{x}_i$,
$m$ denotes an angular margin parameter,
$y_i$ denotes an object label of $\mathbf{x}_i$,
and $s$ denotes a scaling factor that controls smoothness
of the predicted probability distribution.
By setting the angular margin $m$ for the score
of the correct class, the network is trained so that
the intra-class distances of the feature distributions become large
and thus discriminative features are learned.

\subsection{Online learning by the robust weight imprinting}

For online learning of segmentation,
we adopt a few-shot learning based on weight imprinting
\cite{Brown2018}\cite{Siam2019a}.
Suppose we have $N$ pairs of an input image and a training mask
$\{(I_i, M_i)\}^{N}_{i=1}$
and an intermediate feature map $\mathbf{x}_i$ is yielded
by passing the image $I_i$ through the segmentation network.

\begin{figure}[tb]
	\centering
	\includegraphics[width=0.7\hsize]{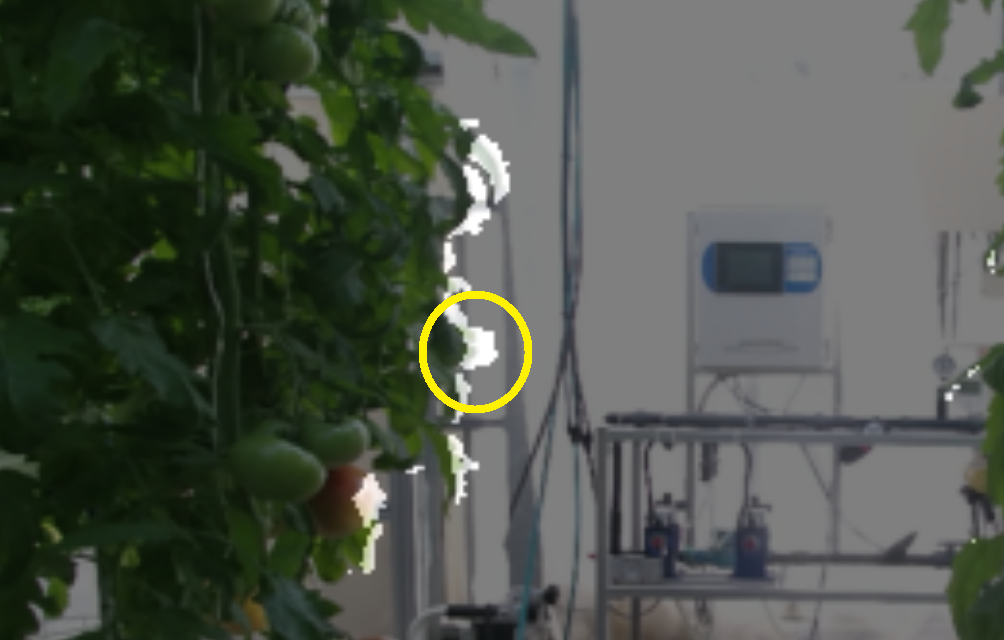}
	\caption{Training mask with error due to noise in the depth sensor}
	\label{fig:mask_error}
\end{figure}

Here, we point out a problem of the training mask
generated by the procedure
described in the previous subsection.
Because of the depth noise and registration error of the
RGB-D sensor,
the training masks include regions set as 1
and not part of the plant regions.
An example of erroneous mask is shown in Fig. \ref{fig:mask_error}.
MAP described in Sec. \ref{sec:weight_imprinting_explanation}
is prone to those outliers
since features are equally averaged,
In our method, we replace it with weighted averaging based on
the distance from the center of the feature distribution within the mask,
which we call \textit{robust average pooling (RAP)}:
\begin{equation}
	\label{eq:x_rap}
	\mathbf{x}_{RAP}=\frac{\sum^{N}_{i=1}\sum^{H}_{h=1}\sum^{W}_{w=1} v^{(h, w)}_iM^{(h,w)}\mathbf{x}^{(h, w)}_i}{\sum^{N}_{i=1}\sum^{H}_{h=1}\sum^{W}_{w=1}M^{(h,w)}},
\end{equation}
where $v^{(h, w)}$ denotes the weight on the corresponding feature $\mathbf{x}^{(h, w)}$
calculated as follows:
\begin{equation}
	v^{(h, w)}_i=\begin{cases}
		\mathbf{x}_i^{(h, w)}\cdot \mathbf{x}_{MAP} & \text{if $\mathbf{x}_i^{(h, w)}\cdot \mathbf{x}_{MAP} \geq 0$} \\
		0                                           & \text{otherwise}.
	\end{cases}
\end{equation}
The averaged feature is then L2-normalized
to form a convolution weight vector for the novel class:
\begin{equation}
	\hat{\mathbf{x}}_{RAP}=\frac{\mathbf{x}_{RAP}}{\|\mathbf{x}_{RAP}\|_2}.
\end{equation}
We directly use $\hat{\mathbf{x}}_{RAP}$ as a weight vector of the classification layer
for the new class representing
false negative plant regions in the model before the training.

%% file: sections/experiments.tex
\section{EXPERIMENTS}

\subsection{Experimental setup}

The DNN model is implemented with PyTorch \cite{Paszke2019}.
The pre-training and inference are performed on one NVIDIA
GeForce GTX 1080Ti with 11GB of memory.
Before the online learning,
we pre-train the scene recognition model in a method
proposed in \cite{Matsuzaki2022},
with modification of the model and the loss function
described in Sec. \ref{sec:network_architecture}.
The angular margin $m$ is set to 0.1.

We use five pairs of an image and a mask
taken in different scenes for few-shot learning.
For testing, we exploit 15 manually labeled images
of scenes where influence of misclassification can be significant,
such as the end of the plant rows,
as reported in \cite{Matsuzaki2022}.

\subsection{Baselines}

As baselines, we adopt weight imprinting with the ordinary
masked average pooling and also a fine-tuning approach via model distillation.
In the former, the averaged feature over the mask (eq. \eqref{eq:x_mean})
is used to calculate the classification weight instead of eq. \eqref{eq:x_rap}.

In model distillation,
we adopt output-level distillation described in \cite{Michieli2021a}.
We set a weight for distillation loss to 0.5 and
temperature parameter $T$ for the softmax function to 1.0.
The entire network is optimized.
We utilize the full mask of interacted regions
since the task is fine-tuning with the existing plant class,
instead of learning a novel class.
The batch size is set to 5 and
a constant learning rate of $1\times 10^{-4}$ is used.
The model is trained for 15 epochs and we report the results
on the epoch with the best mean IoU.

We hereafter denote
the proposed robust weight imprinting as WI-RAP,
the weight imprinting with
the ordinary masked average pooling as WI-MAP,
and the model distillation method as MD.

\subsection{Online model refinement}

Table \ref{table:accuracy_online_learning} shows the comparison of
IoU before and after the online learning
as well as the results of model distillation.
While WI-MAP resulted in degrading the performance,
the proposed robust pooling led to better IoU.
This qualitatively shows the advantage of our method.
MD resulted in better mean IoU and per-class IoU on
the \textit{artificial object} and \textit{ground} classes
than the proposed method.
The proposed method, however, resulted in
better per-class IoU on the \textit{plant} class than MD.
Since our main purpose is to improve the accuracy on
plant regions,
this result shows better suitability of the proposed method
to the online refinement.
\begin{table}
	\centering
	\caption{Per-class and mean IoU before and after the training}
	\label{table:accuracy_online_learning}
	\begin{tabular}{ccccc} \toprule
		                  & Plant          & Artificial obj.   & Ground            & mIoU              \\ \midrule
		Before            & 80.89          & 84.06             & 60.04             & 75.00             \\
		\midrule
		MD                & 80.88          & \textbf{85.90}    & \textbf{63.23}    & \textbf{76.67}    \\
		WI-MAP            & 79.71          & 85.04             & 60.04             & 74.93             \\
		WI-RAP (proposed) & \textbf{81.19} & \underline{85.49} & \underline{60.05} & \underline{75.57} \\ \bottomrule
	\end{tabular}
	\raggedright
	\textbf{Bold} denotes the best and \underline{underline} denotes
	the second best results.
\end{table}

We further look into the precision and the recall metrics.
Table \ref{table:accuracy_online_learning_pr}
shows the precision and the recall of each method.
Our primary purpose is to refine the false negative plant regions
which is relevant to the recall metric.
The both few-shot learning approaches resulted in
better recall of plant regions.
The improvement of the recall, however, came at a cost of
degradation of the precision.
Compared to WI-MAP,
the proposed robust imprinting, WI-RAP, yielded a comparative gain of recall
(-0.92 from WI-MAP)
with less degradation of precision (+2.44 from WI-MAP).
\begin{table}
	\centering
	\caption{Recall and precision before and after the training}
	\label{table:accuracy_online_learning_pr}
	\begin{tabular}{cccc} \toprule
		                  & Plant                     & Artificial obj.           & Ground                                \\ \midrule
		Before            & 89.02 / \textbf{89.86}    & \textbf{92.28} / 90.41    & \underline{93.70} / \underline{62.57} \\
		\midrule
		MD                & 92.21 / \underline{86.80} & \underline{92.03} / 92.81 & \textbf{94.63} / \textbf{65.58}       \\
		WI-MAP            & \textbf{93.77} / 84.16    & 89.72    / \textbf{94.23} & \underline{93.70} / \underline{62.57} \\
		WI-RAP (proposed) & \underline{92.85} / 86.60 & 91.04 / \underline{93.34} & \underline{93.70} / \underline{62.57} \\ \bottomrule
	\end{tabular}
	\raggedright
	The results are shown in the order of recall / precision.

	\textbf{Bold} denotes the best and \underline{underline} denotes
	the second best results.
\end{table}
\begin{figure*}[tb]
	\raggedleft
	\begin{minipage}{0.40\hsize}
		\centering\includegraphics[width=\hsize]{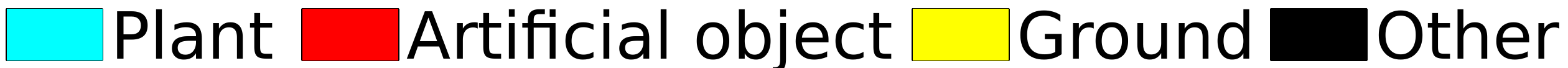}
	\end{minipage}\\

	\begin{minipage}{\hsize}
		\centering
		\includegraphics[width=0.96\hsize]{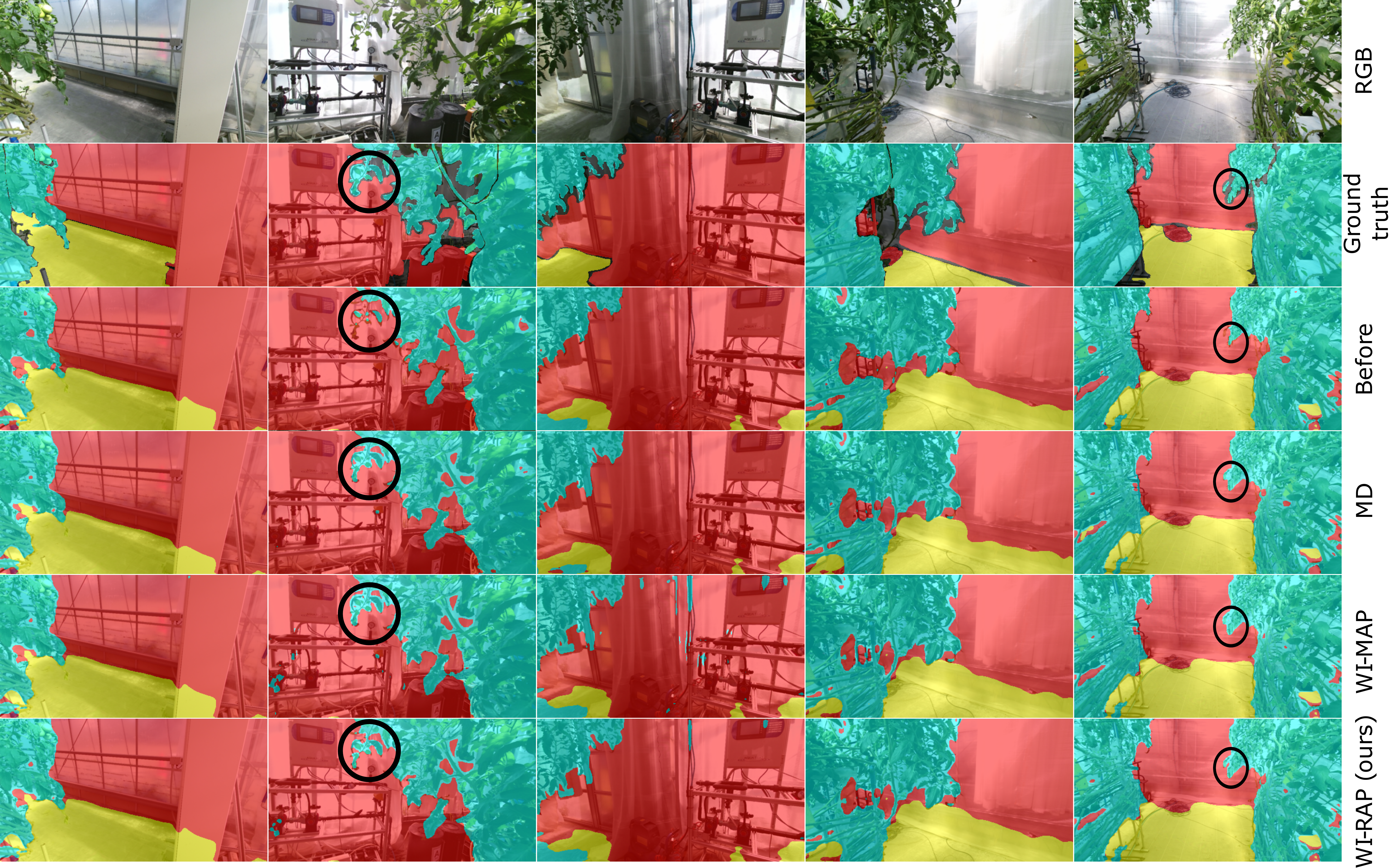}
		\caption{Qualitative evaluation of the online learning.
			After the training, the shapes of plant parts are better captured
			especially in the regions in the black circles.
			Compared to WI-MAP, there are less false positive predictions as \textit{plant} class
			in the predictions of the proposed WI-RAP.
			MD provided similar or better predictions.
			The proposed method, however, resulted in similar results
			with only forward pass unlike MD which requires backpropagation.
		}
		\label{fig:qualitative_online_leanring}
	\end{minipage}
\end{figure*}

Fig. \ref{fig:qualitative_online_leanring} shows the qualitative results.
After the online training by the proposed method,
false negative plant regions are corrected to the \textit{plant} class.
Especially, the regions in black circles show obvious improvement of segmentation.
In those regions, detailed shapes of the leaves and the branches are
better captured compared to the segmentation before training.
Compared to WI-MAP, there are less false positives of prediction as \textit{plant}
in the predictions of the proposed WI-RAP.
This is because while WI-MAP aggressively involves outliers in feature aggregation,
WI-RAP robustly calculate the imprinted weights by the weighted averaging.

One may notice that MD resulted in better performance.
It, however, required around 20 [sec] of training (excluding in-training validation)
and 5 [GB] of GPU memory consumption.
Although the batch size of 1 decreased the memory consumption to about 1.5 [GB],
the results were worse.
In contrast, the proposed method took 0.38 [sec]
and consumed approx. 700 [MB] because it requires
only one forward calculation for each training image.
It is, therefore, more suitable to on-the-fly model refinement on
modern onboard computers with GPU-acceleration such as NVIDIA Jetson.





\subsection{Parameter evaluation}

Next, we evaluate the effect of the angular margin $m$ in
the ArcFace loss.
Table \ref{table:angle_margin} shows the relationship between
$m$ used in model pre-training and the performance of few-shot learning.
\begin{table}
	\centering
	\caption{Mean IoU depending on different angular margin $m$}
	\label{table:angle_margin}
	\begin{tabular}{ccccccc} \toprule
		m      & 0.0   & 0.1   & 0.2   & 0.3   & 0.4   & 0.5   \\ \midrule
		Before & 74.89 & 75.00 & 74.46 & 72.77 & 74.25 & 66.61 \\
		\midrule
		WI-MAP & 41.47 & 74.93 & 68.29 & 72.97 & 74.55 & 66.99 \\
		WI-RAP & 46.80 & 75.57 & 74.96 & 72.76 & 74.09 & 67.00 \\
		\bottomrule
	\end{tabular}
\end{table}
When $m=0.0$, i.e., no angular margin constraint is imposed,
few-shot learning resulted in significant degradation of mean IoU.
This may be because when
the features are learned without the angle margin,
inter-class distances of the features are not encouraged to be larger,
and thus the learned features are not discriminative enough
with large overlaps.
It will lead to confusion of features to be imprinted and those not.
In contrast, when $m=0.1$, the proposed method resulted
in improving the performance,
presumably thanks to better feature representations.
The effectiveness of the weight imprinting, however,
did not hold with all the parameters.
When $m=0.3, 0.4$, the weight imprinting resulted in slight degradation.
The condition of $m$ for effective imprinting is not clear yet.
We leave further analysis of the parameter setting and
the performance for future work.

%

%% file: sections/conclusions.tex
\section{CONCLUSIONS AND FUTURE WORK}

We proposed a novel framework of online model refinement
to deal with misclassification
which may lead the robot to be stuck during navigation.
We introduced a few-shot segmentation method
based on weight imprinting, which allows for
model refinement on the fly. 
Masks for the few-shot training are generated
through observation of a human operator
interacting with plant parts in the environment.
To mitigate the effect of inaccurate masks due to depth noise,
we proposed robust weight imprinting
and showed that the proposed method outperformed
the ordinary weight imprinting with masked average pooling
in terms of IoU.

As future work, we are going to work on more detailed analysis
of the relationship of intermediate feature representation
and the performance of the online learning.
We also consider adopting the proposed method
in the navigation system and conduct real-world experiments.
In addition, designing a better interface for human-robot interaction
to indicate misclassified plant regions more efficiently
is also under consideration.